\documentclass[10pt,twocolumn,letterpaper]{article}

\usepackage[pagenumbers]{cvpr} 

\usepackage[T1]{fontenc}
\usepackage{microtype}
\usepackage{multirow}
\usepackage{placeins}
\usepackage{stfloats}
\newcommand{\IR}{\mathrm{ir}}
\newcommand{\vis}{\mathrm{vis}}

\definecolor{bestpink}{HTML}{D83F87}
\definecolor{secondpurple}{HTML}{9673BD}

\definecolor{cvprblue}{rgb}{0.21,0.49,0.74}
\usepackage[pagebackref,breaklinks,colorlinks,allcolors=cvprblue]{hyperref}
\usepackage[table]{xcolor}

\definecolor{firstcolor}{RGB}{255,182,185}
\definecolor{secondcolor}{RGB}{232,222,242}

\def\confName{CVPR}
\def\confYear{2027}

\title{LIFT: Lifting Language-Controlled Image Fusion to Free-Viewpoint 3D Scenes}

\author{%
	Jingning Tian$^{1,2}$\thanks{Equal contribution} \quad
	Hao Zhang$^{3,4*}$\thanks{Corresponding author} \quad
	Yumin Huang$^{3}$ \quad
	Zizhuo Li$^{3}$ \quad
	Xingyu Jiang$^{5}$ \quad \\
	Erdun Zhao$^{1,2\ensuremath{\dagger}}$ \quad
	Jiayi Ma$^{3,6\ensuremath{\dagger}}$\\
	$^{1}$Hubei Provincial Key Laboratory of Artificial Intelligence and Smart Learning\\
	$^{2}$School of Computer Science, Central China Normal University\\
	$^{3}$Electronic Information School, Wuhan University\\
	$^{4}$Suzhou Institute of Wuhan University\\
	$^{5}$School of Software Engineering, Huazhong University of Science and Technology\\
	$^{6}$School of Robotics, Wuhan University\\
	{\tt\small janine@mails.ccnu.edu.cn, \{zhpersonalbox,jyma2010\}@gmail.com, \{hym1024, zizhuo\_li\}@whu.edu.cn} \\ {\tt\small jiangxy998@hust.edu.cn, erdunz@ccnu.edu.cn}
}

\begin{document}
\maketitle
\begin{abstract}
Multimodal image fusion traditionally generates a single fused image from paired observations at a specific viewpoint, limiting comprehensive scene perception and adaptability to diverse perceptual objectives. To overcome these limitations, we present LIFT, a novel framework for our proposed language-controlled 3D image fusion (L3D-IF) paradigm that extends multimodal image fusion from 2D images to free-viewpoint 3D scenes, enabling language-guided fusion tailored to diverse perceptual objectives. First, LIFT formulates Semantic Modality Binding (SMB), which interprets language instructions to identify semantic targets and their associated modality preferences and grounds these targets in corresponding regions of a shared Gaussian scene, establishing persistent semantic-to-spatial correspondences for consistent fusion control across viewpoints. Second, Language-Conditioned Gaussian Appearance (LCGA) is developed to achieve scene-wide multimodal appearance modeling by combining adaptive multimodal blending with language- and viewpoint-conditioned appearance refinement, enabling spatially adaptive realization of diverse fusion appearances while preserving source fidelity at pure-modality endpoints. Finally, we construct a Multi-View Language-Supervision (MVLS) dataset that associates language instructions and calibrated viewpoints with modulated fused images, providing effective supervision for learning instruction-dependent fusion from finite multi-view observations. Extensive experiments demonstrate that LIFT achieves coherent multimodal fusion across continuous viewpoints, accurate semantic target localization, and consistent regional fusion control under diverse language instructions. Moreover, LIFT enables previously observed targets to remain semantically accessible even when occluded from the current viewpoint, highlighting its advantage in maintaining scene-level semantic awareness beyond individual viewpoints.
\end{abstract}

\section{Introduction}
\label{sec:intro}
Images acquired through a single sensing modality reveal only a partial account of the physical world. For example, visible cameras capture color and fine textures, whereas infrared cameras expose thermal distribution. Multimodal image fusion (MMIF) aims to combine these complementary observations into a more informative representation, supporting a wide range of applications, \textit{e.g.}, intelligent security and assisted driving~\cite{liu2025infrared,yi2026ll,xiang2026single}. Yet most existing image fusion methods formulate this objective in two-dimensional (2D) space, using an aligned image pair to generate a single fused image from a specific viewpoint~\cite{zhang2021image,xu2022u2fusion,zhang2026dpafuse}. Under this \textbf{\textit{2D image fusion (2D-IF)}} paradigm, the relationships among multimodal visual cues are encoded in a transient pixel arrangement tied to the captured viewpoint, rather than preserved as persistent properties of the imaging scene.

\begin{figure*}[t]\centering
\includegraphics[width=0.95\linewidth]{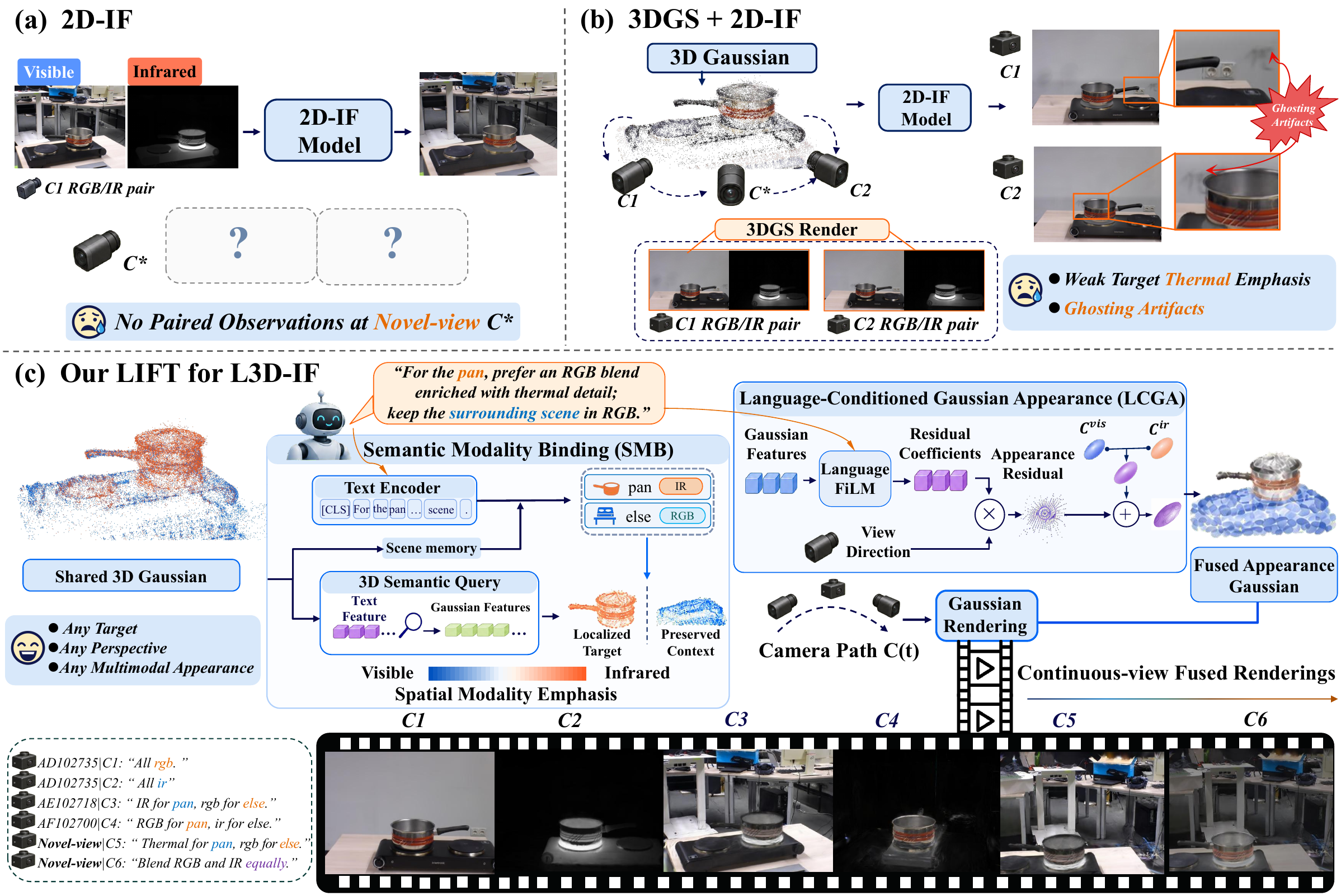}
\vspace{-0.2cm}
\caption{A paradigm shift from 2D image fusion to 3D image fusion. (a) 2D-IF: conventional fusion of paired multimodal images from a given viewpoint. (b) 3DGS+2D-IF: 3DGS-based multi-view rendering followed by 2D fusion. (c) Our LIFT for L3D-IF: our proposed language-controlled fusion paradigm within a unified 3D scene representation.}
\label{fig:intro}
\end{figure*}

More fundamentally, the view-dependent nature of the current 2D-IF paradigm limits its ability to represent a scene as a whole. A fused image captures only the multimodal information visible from a particular camera pose, leaving occluded surfaces, out-of-view regions, and unseen sides of objects unrepresented. Simply extending 2D-IF models to multiple discrete viewpoints does not resolve this limitation, as independently fused images lack explicit spatial connections and cannot form a unified scene representation. Moreover, paired source observations at novel viewpoints are not readily available, as shown in Fig.~\ref{fig:intro} (a). Recent advances in three-dimensional (3D) scene representations, particularly 3D Gaussian Splatting (3DGS), offer a promising direction for overcoming these limitations by enabling free-viewpoint multimodal scene reconstruction~\cite{kerbl20233d,lu2025thermalgaussian}. Building on this capability, a straightforward solution is to first use 3D scene representations to render modality-specific observations from arbitrary viewpoints, and then apply conventional 2D fusion methods to the rendered image pairs. Nevertheless, this cascaded strategy still performs fusion independently at each viewpoint, without explicitly integrating the fused information into a unified 3D scene representation. As a result, the same scene regions may exhibit inconsistent fusion results across different viewpoints, making it difficult to achieve comprehensive and consistent scene perception, as shown in Fig.~\ref{fig:intro} (b). This raises an appealing prospect: moving image fusion from independently processed 2D observations into the 3D scene representation itself, giving rise to a new paradigm of \textbf{\textit{3D Image Fusion (3D-IF)}}.

Beyond constructing a unified 3D scene representation, an important consideration is how 3D-IF can serve diverse perceptual objectives. Different tasks may require substantially different visual information from the same scene. For example, when a robot cleans a cooking pot, visible cues provide more informative details for identifying residual stains and assessing surface conditions, whereas when using the same pot for cooking, infrared cues become more important for perceiving temperature distributions and monitoring its heating state. This illustrates that the perceptual relevance of each modality depends not only on the semantic content of the scene but also on the intended task. Therefore, a fixed multimodal mixture cannot adequately accommodate the varying information requirements of diverse perceptual objectives. In this context, 3D-IF should support the dynamic adaptation of fusion behaviors during inference. Language provides a natural and flexible means of expressing high-level perceptual intentions, offering an intuitive interface for guiding such adaptations. These considerations motivate a language-controlled 3D image fusion (L3D-IF) paradigm.

However, realizing L3D-IF is nontrivial and entails two fundamental challenges. First, language instructions describe semantic targets and their intended multimodal appearances, whereas a 3D scene consists of spatial primitives without explicit semantic correspondences. This requires identifying the intended semantic regions and establishing their spatial correspondences within the 3D scene. Such correspondences must persist across viewpoints, ensuring that the same scene content can be consistently identified regardless of the viewing direction. Second, even when the intended target regions are correctly associated with their corresponding 3D locations, their desired multimodal appearances are not necessarily guaranteed. As the viewpoint changes, the visible portions of a target and the available information from different modalities may vary, making it challenging to preserve the intended fusion characteristics in the rendered results. Moreover, reference renderings of the desired multimodal appearances are unavailable across diverse language instructions and viewpoints, limiting the learning of L3D-IF.

To address these challenges, we propose \textbf{LIFT}, a novel framework for L3D-IF that extends image fusion from viewpoint-specific 2D images to unified 3D scene representations. First, to establish persistent semantic-to-spatial correspondences, we introduce Semantic Modality Binding (SMB), which grounds language instructions in spatially consistent regions of a shared Gaussian scene. Specifically, a scene-conditioned controller identifies semantic targets and their modality preferences, then queries the semantic field to assign continuous fusion weights to the corresponding Gaussian primitives. By anchoring these assignments to scene content rather than individual image pixels, SMB enables consistent semantic localization and modality selection across viewpoints. Second, building on these correspondences, we further develop Language-Conditioned Gaussian Appearance (LCGA) to realize instruction-specific multimodal appearances within the 3D scene. LCGA combines visible and infrared source colors according to the assigned modality preferences and predicts residual appearance corrections conditioned on language and viewing direction, thereby adapting regional appearances to perceptual intentions while preserving source colors at exact modality endpoints. Finally, to address the lack of reference renderings for diverse language instructions and viewpoints, we construct a Multi-View Language-Supervision (MVLS) dataset for learning language-guided continuous-view fusion. Each sample associates a language instruction and a calibrated camera pose with a modulated fused image. These samples guide LIFT to learn instruction-dependent fusion behaviors from finite multi-view observations, enabling coherent fusion across continuous viewpoints. Together, these components enable our LIFT to achieve comprehensive 3D scene perception while accommodating diverse perceptual objectives through language control, as shown in Fig.~\ref{fig:intro} (c).

In summary, our contributions are as follows:
\begin{itemize}
\item We propose LIFT, a novel framework for the proposed L3D-IF paradigm, advancing multimodal image fusion from viewpoint-specific 2D image generation to language-controlled 3D scene perception, with the flexibility to accommodate diverse perceptual objectives.
\item We develop Semantic Modality Binding (SMB) to establish persistent semantic-to-spatial correspondences between language instructions and 3D Gaussian primitives, enabling consistent semantic localization and modality selection across viewpoints.
\item Language-Conditioned Gaussian Appearance (LCGA) is developed to realize instruction-specific multimodal appearances in 3D scenes by combining modality-dependent source colors with language- and viewpoint-conditioned residual corrections, while preserving source fidelity at exact modality endpoints.
\item We construct a Multi-View Language-Supervision (MVLS) dataset that associates language instructions and calibrated viewpoints with modulated fused images, providing supervision for learning instruction-dependent fusion from finite multi-view observations while supporting future research on L3D-IF.
\end{itemize}

\begin{figure*}[t]
	\centering
	\includegraphics[width=0.95\linewidth]{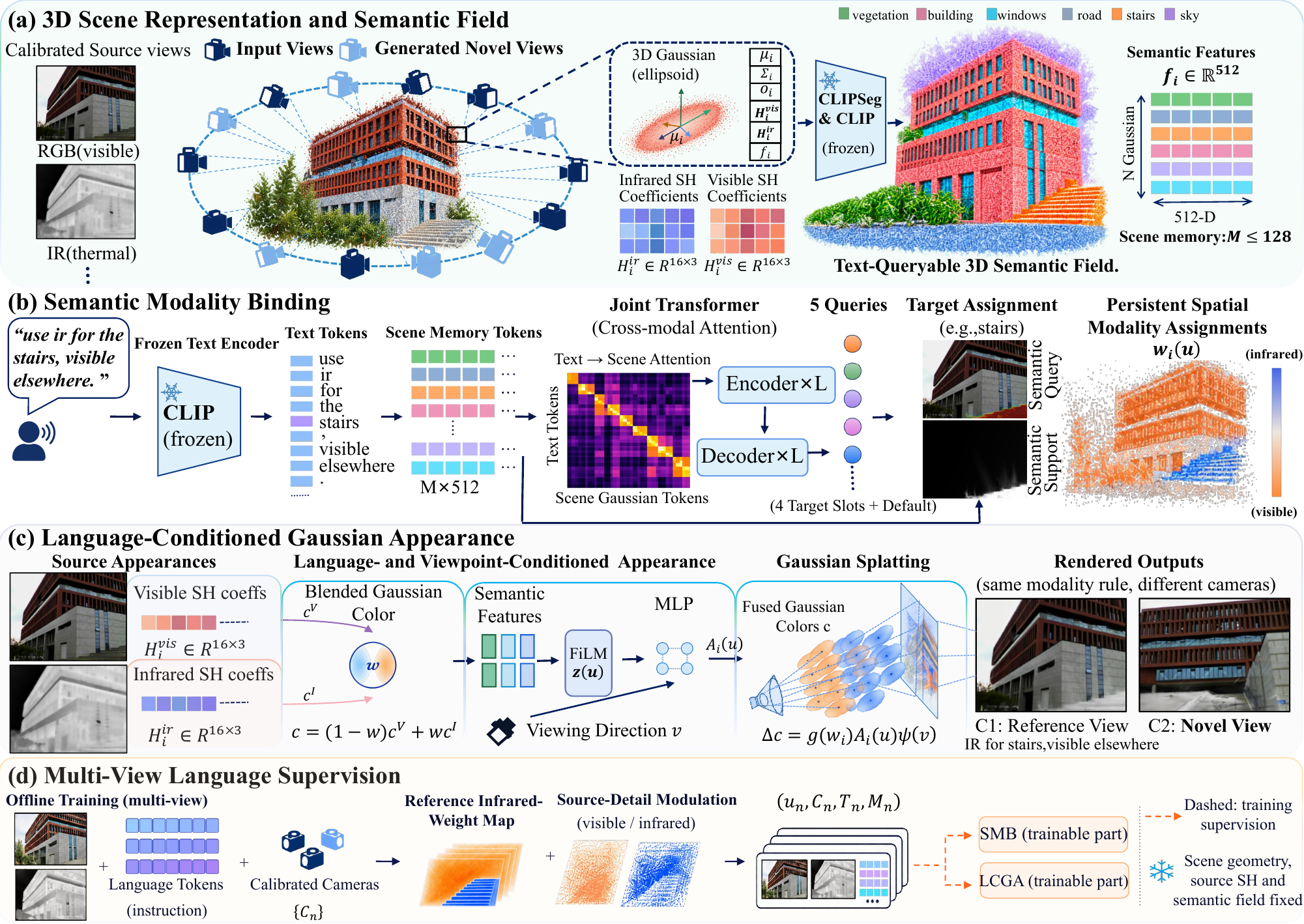}
	\vspace{-0.2cm}
	\caption{The overall pipeline of our LIFT. (a) Shared scene reconstruction: OMMG provides shared geometry and bimodal appearances, while frozen CLIP/CLIPSeg supervise the semantic field. (b) SMB translates language instructions into persistent per-Gaussian modality assignments. (c) LCGA blends source appearances and applies language- and viewpoint-conditioned residual refinement before Gaussian splatting. (d) MVLS constructs multi-view language-supervised targets for SMB pretraining and joint optimization of SMB and LCGA.}
	\label{fig:method}
\end{figure*}

\section{Methodology}\label{sec:methodology}
LIFT establishes a language-controlled multimodal fusion framework within a unified 3D Gaussian scene representation, as illustrated in Fig.~\ref{fig:method}. It comprises three main components: 3D Scene Representation and Semantic Field, Semantic Modality Binding (SMB), and Language-Conditioned Gaussian Appearance (LCGA). We further construct the Multi-View Language-Supervision (MVLS) dataset to supervise the optimization of SMB and LCGA. The following sections detail these components.

\subsection{3D Scene Representation and Semantic Field}\label{sec:scene}
\noindent\textbf{Bimodal Gaussian Scene Reconstruction.}
We adopt ThermalGaussian's one multi-modal gaussian (OMMG) representation~\cite{lu2025thermalgaussian}. Each of the $N$ Gaussian primitives is characterized by a shared center $\boldsymbol\mu_i\in\mathbb R^3$, positive-definite covariance $\boldsymbol\Sigma_i\in\mathbb R^{3\times3}$, and opacity $o_i\in[0,1]$, together with separate degree-three spherical-harmonic (SH) coefficients for visible and infrared appearances, denoted by $\mathbf H_i^{\vis},\mathbf H_i^{\IR}\in\mathbb R^{16\times3}$~\cite{kweon2025mrgs}.

Given calibrated pairs $(V_j,I_j,C_j)$, where $V_j$ and $I_j$ denote registered visible and infrared images and $C_j$ denotes the corresponding camera, we reconstruct the shared 3D geometry and bimodal appearances. All input images have three channels with values in $[0,1]$, with infrared intensities replicated after a fixed display conversion. Starting from sparse reconstruction, we first optimize the geometry, opacity, and visible SH coefficients through RGB fitting, incorporating Gaussian densification and pruning. Then, we fit the infrared SH coefficients while keeping the shared geometry fixed. Both stages employ a combination of mean $\mathcal L_1$ and structural similarity (SSIM) reconstruction losses.

For a camera center $\mathbf o_C$, the viewing direction of Gaussian $i$ relative to this camera position is defined as:
\begin{equation}
	\mathbf v_i(C)=
	\frac{\boldsymbol\mu_i-\mathbf o_C}
	{\|\boldsymbol\mu_i-\mathbf o_C\|_2}.
\end{equation}
Evaluating the corresponding SH functions with the standard $0.5$ offset and nonnegative clamp~\cite{kerbl20233d} yields the view-dependent colors $\mathbf c_i^{\vis}(C),\mathbf c_i^{\IR}(C)\in\mathbb R_{\geq0}^3$. Both modalities share the same geometric contributions $\rho_i^C(\mathbf p)$, defined as the product of the projected alpha and the accumulated transmittance of preceding Gaussians at pixel $\mathbf p$. The remaining transmittance determines the background contribution.

\noindent\textbf{Text-Queryable 3D Semantic Field.}
To establish semantic correspondences between language descriptions and Gaussian primitives, we associate each Gaussian with a learnable semantic feature $\mathbf f_i\in\mathbb R^{512}$. Frozen CLIP~\cite{radford2021clip} provides unit-normalized phrase embeddings $\mathbf e(t)\in\mathbb R^{512}$ and dense image features $\mathbf F_j$ extracted from projected and normalized patch tokens. Meanwhile, CLIPSeg~\cite{luddecke2022segmentation} provides phrase-conditioned segmentation probabilities $S_j(\mathbf p;t)\in[0,1]$, which are replaced by annotations where available. Given a semantic phrase $t$, its response on Gaussian $i$ is computed as:
\begin{equation}
	q_i(t)=\sigma\!\left[
	\tau\left(
	\frac{\mathbf f_i^{\mathsf T}\mathbf e(t)}
	{\max(\|\mathbf f_i\|_2,\varepsilon_f)}-b
	\right)\right],
	\label{eq:query-main}
\end{equation}
where $\sigma$ denotes the sigmoid, $\ell$ and $b$ are learnable calibration parameters, $\tau=\operatorname{clip}_{[5,80]}(e^\ell)$ is an inverse-temperature scale, and $\varepsilon_f=10^{-12}$ stabilizes feature normalization.

To connect image-level supervision with the shared 3D semantic features~\cite{qin2024langsplat}, we project Gaussian features and phrase-query responses onto the observed views. Specifically, given the accumulated Gaussian coverage $\mathcal A_j(\mathbf p)=\sum_i\rho_i^{C_j}(\mathbf p)$, the rendered semantic features and segmentation probabilities are computed as:
\begin{equation}
	\begin{aligned}
		\widehat{\mathbf F}_j(\mathbf p)
		&=\frac{\sum_i\rho_i^{C_j}(\mathbf p)\mathbf f_i}
		{\max(\mathcal A_j(\mathbf p),10^{-6})},\\
		\widehat S_j(\mathbf p;t)
		&=\frac{\sum_i\rho_i^{C_j}(\mathbf p)q_i(t)}
		{\max(\mathcal A_j(\mathbf p),10^{-6})},
	\end{aligned}
	\label{eq:semantic-projection}
\end{equation}
here, $\widehat{\mathbf F}_j(\mathbf p)\in\mathbb R^{512}$ is not unit-normalized, while $\widehat S_j(\mathbf p;t)\in[0,1]$ represents the rendered semantic response. We optimize the Gaussian semantic features $\{\mathbf f_i\}$ and calibration parameters $\ell,b$ using confidence-weighted binary cross-entropy (BCE) for segmentation masks and mean squared error (MSE) for dense features. The confidence weights exclude ambiguous teacher probabilities and poorly covered pixels. Supervising the same Gaussian features across views, consolidates semantic evidence into a shared 3D representation.

\noindent\textbf{Evidence-Supported Queries.}
To incorporate semantic evidence from the training RGB views
$\mathcal T$, we optionally aggregate CLIPSeg probabilities
using Gaussian rendering contributions and cache the resulting
per-Gaussian semantic support. Specifically, defining the total contribution mass of Gaussian $i$ as $\nu_i=\sum_{j\in\mathcal T}\sum_{\mathbf p}\rho_i^{C_j}(\mathbf p)$, its cached semantic support is computed as:
\begin{equation}
	q_i^{\rm ev}(t)=
	\begin{cases}
		\displaystyle
		\frac{
			\sum_{j\in\mathcal T}\sum_{\mathbf p}
			\rho_i^{C_j}(\mathbf p)S_j(\mathbf p;t)
		}{\nu_i}, & \nu_i>0,\\[2.2ex]
		0, & \nu_i=0.
	\end{cases}
	\label{eq:evidence-query}
\end{equation}
The effective query response $q_i^{\rm eff}(t)$ uses Eq.~\eqref{eq:evidence-query} for phrases with enabled evidence caching and Eq.~\eqref{eq:query-main} otherwise. For distilled queries, responses are set to zero on Gaussians that were not observed during semantic-field fitting. Continuous query scores are retained for fusion, whereas target-only projection selects Gaussians satisfying $q_i^{\rm eff}(t)\geq0.5$ and preserves their original opacities.

\subsection{Semantic Modality Binding}\label{sec:smb}
\noindent\textbf{Scene-Conditioned Instruction Decoding.}
SMB takes frozen CLIP instruction tokens together with $N_{\rm mem}\leq128$ sampled semantic features as inputs. After applying learned projections and type embeddings, these features are fed into a Transformer encoder-decoder. Specifically, $K=4$ queries predict ordered phrase-infrared-ratio slots $(t_k,a_k)$, where $t_k$ denotes a semantic phrase and $a_k\in[0,1]$ specifies its infrared ratio, while a fifth query predicts the default infrared ratio $d\in[0,1]$. Mode heads further distinguish visible ($a_k=0$), infrared ($a_k=1$), mixed, and inactive slots.

To extract structured control information from the instruction, span heads identify semantic phrases, while a numeric pointer binds explicitly specified percentages. For qualitative modality preferences, a sigmoid ratio head predicts the corresponding infrared ratios with ordinal adaptation. The mean query states provide the LCGA conditioning context $\mathbf z(u)\in\mathbb R^{256}$, while a direction head assists camera selection.

\noindent\textbf{Regional Modality Assignment.}
Given the decoded instruction slots, SMB grounds each active semantic phrase in the 3D scene using the effective query responses established in Sec.~\ref{sec:scene}. Specifically, for an active slot $k$, its semantic support on Gaussian $i$ is given by $q_{ik}=q_i^{\rm eff}(t_k)$. For inactive slots, we set $q_{ik}=0$ and the corresponding unused ratio $a_k=0$. Starting from the default infrared ratio $w_i^{(0)}=d$, SMB sequentially applies the ordered rules to update the Gaussian-level infrared weights:
\begin{equation}
	\begin{aligned}
		w_i^{(k)}
		&=(1-q_{ik})w_i^{(k-1)}+q_{ik}a_k,\\
		w_i(u)&=w_i^{(K)},\qquad k=1,\ldots,K,
	\end{aligned}
	\label{eq:composition-main}
\end{equation}
where $q_{ik}$ determines the semantic support of rule $k$ on Gaussian $i$, while $a_k$ specifies its target infrared ratio. The resulting convex updates ensure that $w_i(u)\in[0,1]$ and resolve overlapping rules according to their prescribed order. Later rules fully override earlier assignments only when their semantic support reaches unity, while the default ratio is retained where all semantic supports vanish. Hence, uncertain semantic support may prevent exact modality endpoints. Importantly, the resulting weights remain attached to their corresponding Gaussian primitives across different viewpoints, providing persistent spatially grounded modality assignments for subsequent appearance modeling.

\subsection{Language-Conditioned Gaussian Appearance}\label{sec:lcga}
\noindent\textbf{Language- and Viewpoint-Conditioned Appearance.}
LCGA combines a 64-dimensional projection of the Gaussian semantic feature $\mathbf f_i$ with the language context $\mathbf z(u)$ through feature-wise linear modulation (FiLM)~\cite{perez2018film} and an MLP. It predicts a residual appearance matrix $\mathbf A_i(u)\in\mathbb R^{3\times4}$, with each coefficient bounded within $[-0.05,0.05]$ using a scaled $\tanh$ function. The output layer is initialized to zero. To incorporate viewpoint-dependent appearance variations, we define the directional encoding as:
\begin{equation}
	\boldsymbol\psi(\mathbf v_i(C))
	=
	[1,v_x,v_y,v_z]^{\mathsf T}
	\in\mathbb R^4,
\end{equation}
where $\mathbf v_i(C)=[v_x,v_y,v_z]^{\mathsf T}$ denotes the viewing direction of Gaussian $i$ under camera $C$, as defined in Sec.~\ref{sec:scene}. Meanwhile, we introduce a modality-dependent gating function $g(w)=4w(1-w)$ to regulate the residual contribution. For brevity, let $w_i=w_i(u)$ denote the infrared weight assigned to Gaussian $i$ by SMB. The resulting instruction-conditioned Gaussian color is computed as:
\begin{equation}
	\begin{aligned}
		\mathbf c_i^{u,C}=\Big[
		&(1-w_i)\mathbf c_i^{\vis}(C)
		+w_i\mathbf c_i^{\IR}(C)\\
		&+g(w_i)\mathbf A_i(u)
		\boldsymbol\psi(\mathbf v_i(C))
		\Big]_+,
	\end{aligned}
	\label{eq:color-main}
\end{equation}
where $[\cdot]_+$ clamps negative color components to zero. The first two terms perform continuous blending of visible and infrared source colors, while the last term provides language- and viewpoint-conditioned residual refinement. Notably, since $g(w)=0$ at $w\in\{0,1\}$, the residual correction vanishes at exact modality endpoints, preserving the corresponding source colors.

\noindent\textbf{Fusion Rendering.}
Given the instruction-conditioned Gaussian colors, the final fused image is obtained by Gaussian splatting with the shared geometric contributions $\rho_i^C(\mathbf p)$:
\begin{equation}
	\widehat \xi(u,C)=
	\operatorname{Splat}
	\left(
	\{\mathbf c_i^{u,C}\},
	\{\rho_i^C\}
	\right),
\end{equation}
where $\operatorname{Splat}(\cdot)$ denotes Gaussian splatting with alpha compositing, and $\widehat \xi(u,C)$ represents the rendered fused image under instruction $u$ and camera $C$. Rendering is performed over a black background or a fixed bimodal sky. For the latter, sky colors are determined using the mean modality weight of a fixed set of sky-associated Gaussians. Image-level endpoint preservation additionally requires the corresponding source background. Along a camera path, SMB's modality weights remain fixed, while LCGA updates the directional appearance and Gaussian splatting resolves visibility changes. When a new instruction is provided, the modality weights and language context are updated accordingly.

\subsection{Multi-View Language-Supervision}\label{sec:mvls}
To supervise language-controlled fusion across diverse viewpoints, we construct an MVLS dataset associating language instructions and calibrated camera poses with source-derived fused targets. We then use MVLS to train SMB and LCGA.

\noindent\textbf{MVLS Construction.}
MVLS is constructed using selected scenes from three existing datasets: RGBT-Scenes~\cite{lu2025thermalgaussian}, ThermalMix~\cite{ozer2024exploring}, and ThermoScenes~\cite{hassan2025thermonerf}. More concretely, MVLS contains $79,854$ instruction-image-camera triplets across $15$ scenes, constructed from $32,414$ distinct view-rule cases at $1,526$ calibrated viewpoints. Different instruction paraphrases may share the same target image and camera. Each sample associates a language instruction $u_n$, a source-derived fused target $T_n$, and a calibrated camera $C_n$, together with a confidence map $M_n$ and structured rule labels.

We construct the fused targets offline from registered visible-infrared pairs $V_n,I_n\in[0,1]^{H_n\times W_n\times3}$. Specifically, applying the ordered modality assignment rules in Eq.~\eqref{eq:composition-main} to CLIPSeg-predicted pixel-wise target probability maps yields a reference infrared-weight map $\widetilde w_n\in[0,1]^{H_n\times W_n}$. Based on this map, we generate candidate fused targets:
\begin{equation}
	\begin{aligned}
		T_{n,m}=\operatorname{Clamp}_{[0,1]}\big[
		&(1-\widetilde w_n)\odot V_n
		+\widetilde w_n\odot I_n\\
		&+g(\widetilde w_n)\odot D_{n,m}
		\big],
	\end{aligned}
	\label{eq:reference-main}
\end{equation}
where $m$ indexes a predefined candidate strategy for generating fused targets, and $D_{n,m}$ denotes its source-detail correction, with the same dimensions as the input images, $\operatorname{Clamp}_{[0,1]}$ denotes element-wise clamping to $[0,1]$. To structure $D_{n,m}$, five candidate strategies are considered, \textit{i.e.}, linear blending, multiscale activity-based fusion, maximum-detail selection, weighted detail enhancement, and CDDFuse-guided correction~\cite{zhang2023cddfuse}. The operator $\odot$ denotes element-wise multiplication. The gating function $g$ preserves the pure-modality endpoints. We select one strategy $m^\star$ for each scene using its training views and keep this selection fixed across data splits, yielding $T_n=T_{n,m^\star}$. Linear mixing, corresponding to $D_{n,m}=0$, serves as the fallback strategy. The confidence maps suppress ambiguous teacher regions, excluding samples with unreliable semantic support.

\noindent\textbf{Appearance and Control Supervision.}
After scene preparation, SMB is pretrained using structured instruction supervision and then jointly optimized with LCGA. Specifically, MVLS provides appearance and local-structure supervision, while structured labels supervise instruction binding and residual regularization constrains deviations from source-color mixing. The overall objective is defined as:
\begin{equation}
	\begin{aligned}
		\mathcal L_{\rm joint}
		=\mathbb E_{\mathcal D_{\rm train}}\big[
		&\mathcal L_{\rm app}
		+\lambda_1\mathcal L_{\rm grad}\\
		&+\lambda_2\mathcal L_{\rm SMB}
		+\lambda_3\mathcal L_{\rm res}
		\big],
	\end{aligned}
	\label{eq:loss-main}
\end{equation}
where $\mathcal D_{\rm train}$ denotes the training distribution, and $\widehat \xi=\widehat \xi(u,C)$ represents the rendered fused image. The appearance and gradient losses are defined as: $\mathcal L_{\rm app}=\|\widehat \xi-T\|_{M,1}$, $\mathcal L_{\rm grad}=\|\nabla\widehat \xi-\nabla T\|_{M,1}$, where $T$ and $M$ denote the corresponding fused target and confidence map. The confidence-weighted norm averages absolute errors over image channels and confident pixel mass, using $C_X\max(\sum_{\mathbf p}M(\mathbf p),1)$ as the denominator, where $C_X=3$ for appearance and $C_X=6$ for gradients. Horizontal and vertical forward differences produce the six gradient channels. The structured supervision loss $\mathcal L_{\rm SMB}$ supervises semantic spans, modality modes, infrared ratios, default ratios, viewing directions, and enabled numeric pointers. The residual regularization loss is defined as $\mathcal L_{\rm res}=\frac{1}{3N}\sum_i\|\Delta\mathbf c_i\|_2^2$, where $\Delta\mathbf c_i$ denotes the instruction-conditioned Gaussian color in Eq.~\eqref{eq:color-main}, penalizing excessive deviations from the source-color mixture. We set $\lambda_1=0.2$, $\lambda_2=0.05$, and $\lambda_3=0.001$ for learning balance.

During training, annotated candidate-phrase supports are combined differentiably to propagate image-level gradients to SMB, while inference uses the decoded semantic spans. The scene representation and pretrained encoders remain fixed during optimization, with each scene fitted independently without online teacher segmentation.

\begin{figure*}[t]
	\centering
	\includegraphics[width=0.95\linewidth]{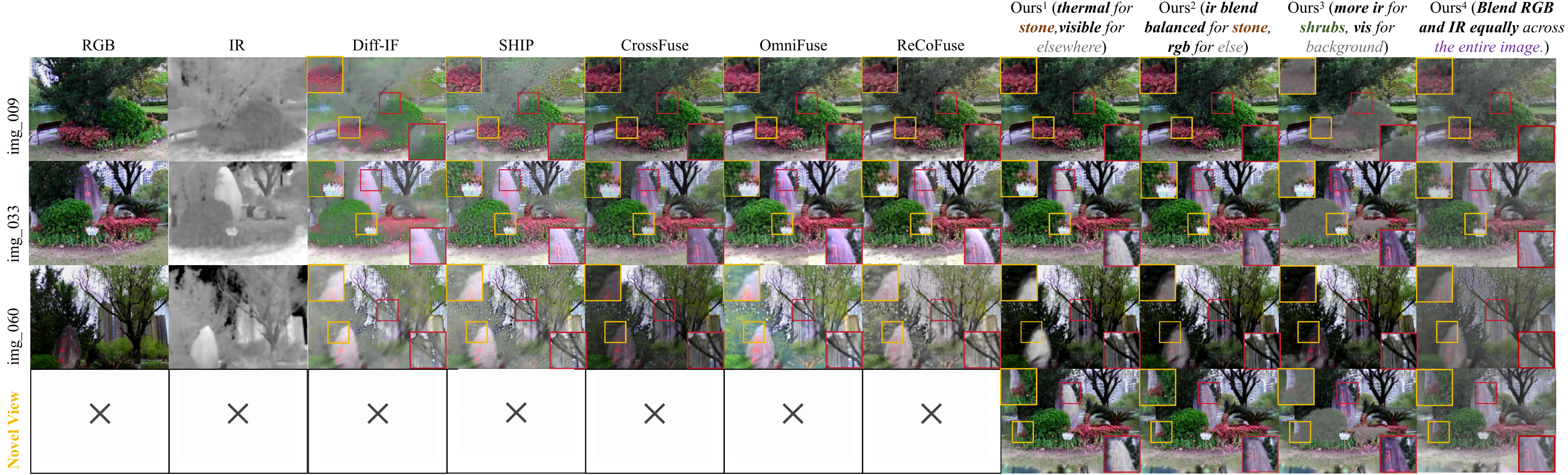}\\[-1pt]
	{\small (a)  Comparisons with 2D-IF methods on the LandScape scene.}\\[3pt]
	\vspace{-0.1cm}
	\includegraphics[width=0.95\linewidth]{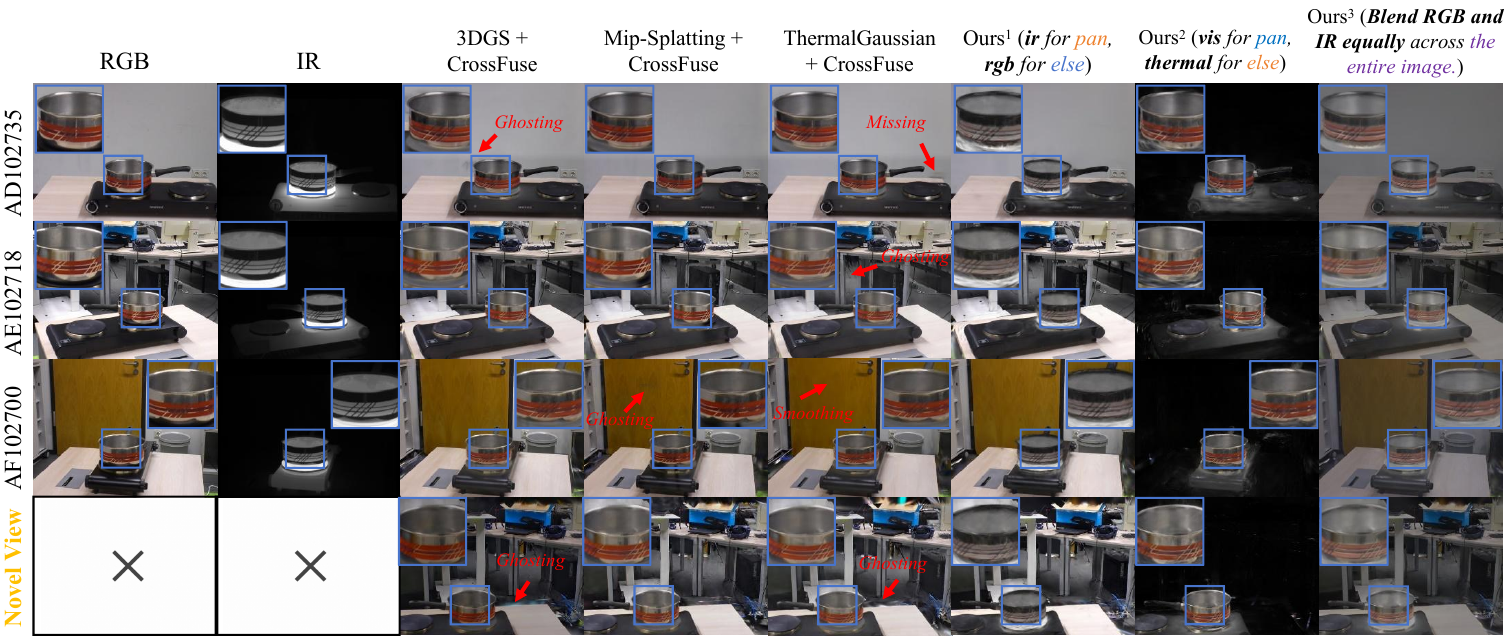}\\[-1pt]
	{\small (b) Comparisons with 3DGS+2D-IF methods on the Pan Scene.}
	\vspace{-0.2cm}
	\caption{Free-viewpoint fusion comparisons. Each scene includes three captured views and one novel view.}
	\label{fig:main}
	\vspace{-0.1cm}
\end{figure*}

\section{Experiments} \label{sec:experiments}
\subsection{Experimental Setup}
The evaluation covers $15$ scenes from the constructed MVLS dataset. Fusion quality is evaluated using mean squared error (MSE), peak signal-to-noise ratio (PSNR), correlation coefficient (CC), and the fusion artifact metric $N^{AB/F}$. We further introduce \textit{Regional MAE}, \textit{Reproj. MAE} to evaluate spatial regional modality accuracy . Specifically, \textit{Regional MAE} measures deviations between predicted and instructed infrared weights in manually annotated target and background regions. \textit{Reproj. MAE} quantifies cross-view weight inconsistencies at corresponding pixels. All experiments are conducted on NVIDIA GeForce RTX 4090 GPUs and an Intel Xeon Platinum 8352V CPU.

\subsection{Language-Controlled Free-Viewpoint Fusion} \label{fusion apperance}
We compare LIFT with five 2D fusion methods on the LandScape scene, including Diff-IF~\cite{yi2024diffif}, SHIP~\cite{zheng2024ship}, CrossFuse~\cite{li2024crossfuse}, OmniFuse~\cite{zhang2025omnifuse}, and ReCoFuse~\cite{zhang2026recofuse}. On the Pan scene, we further compare against three cascaded 3DGS+2D-IF pipelines, combining 3DGS~\cite{kerbl20233d}, Mip-Splatting~\cite{yu2024mipsplatting}, or ThermalGaussian~\cite{lu2025thermalgaussian} with CrossFuse.

As shown in Fig.~\ref{fig:main}, 2D fusion methods cannot generate fused results at novel viewpoints, while cascaded 3DGS+2D-IF pipelines exhibit ghosting artifacts and cross-view inconsistencies due to independent viewpoint-wise fusion. In contrast, our LIFT grounds modality assignments in a unified 3D scene, enabling consistent language-guided fusion of arbitrary target regions across captured and novel viewpoints. This allows users to immersively explore the scene and perceive its multimodal properties from different viewpoints. For quantitative evaluation, LIFT is tested under the mixed instruction ``\textit{Blend RGB and IR equally across the entire image}'' and compared with competitors across all $15$ scenes. Table~\ref{tab:fusion-main} reports the average results across these scenes. LIFT achieves the best scores on most metrics, indicating its better representation capability.

\begin{table}[!ht]
	\centering
	\caption{Quantitative comparison on $217$ views from $15$ scenes.}\label{tab:fusion-main}
	\vspace{-0.3cm}
	\footnotesize
	\setlength{\tabcolsep}{1pt}
	\renewcommand{\arraystretch}{0.8}
	\begin{tabular*}{\columnwidth}{@{\extracolsep{\fill}}lcccc@{}}
		\toprule
		Method & MSE $\downarrow$ & PSNR $\uparrow$ & CC $\uparrow$ & $N^{AB/F}\downarrow$ \\
		\midrule
		Diff-IF   & 0.086 & 11.117 & 0.493 & 0.011 \\
		SHIP      & 0.084 & 11.310 & 0.485 & 0.020 \\
		CrossFuse & 0.075 & 11.909 & 0.453 & \cellcolor{firstcolor}\textbf{0.003} \\
		OmniFuse  & 0.080 & 11.470 & 0.494 & 0.011 \\
		ReCoFuse  & \cellcolor{secondcolor}\underline{0.072} & \cellcolor{secondcolor}\underline{11.979} & \cellcolor{secondcolor}\underline{0.513} & 0.015 \\
		\midrule
		3DGS + CrossF.           & 0.076 & 11.670 & 0.386 & 0.019 \\
		Mip-Splat. + CrossF.   & 0.079 & 11.515 & 0.370 & 0.019 \\
		ThermalG. + CrossF. & 0.078 & 11.615 & 0.400 & 0.018 \\
		\midrule
		\textbf{Ours} &\cellcolor{firstcolor} \textbf{0.049} & \cellcolor{firstcolor}\textbf{13.542} & \cellcolor{firstcolor}\textbf{0.524} &\cellcolor{secondcolor} \underline{0.005} \\
		\bottomrule
	\end{tabular*}
	\vspace{-0.5cm}
\end{table}

\subsection{Persistent Regional Control}
\begin{figure}[t]\centering
	\includegraphics[width=1\linewidth]{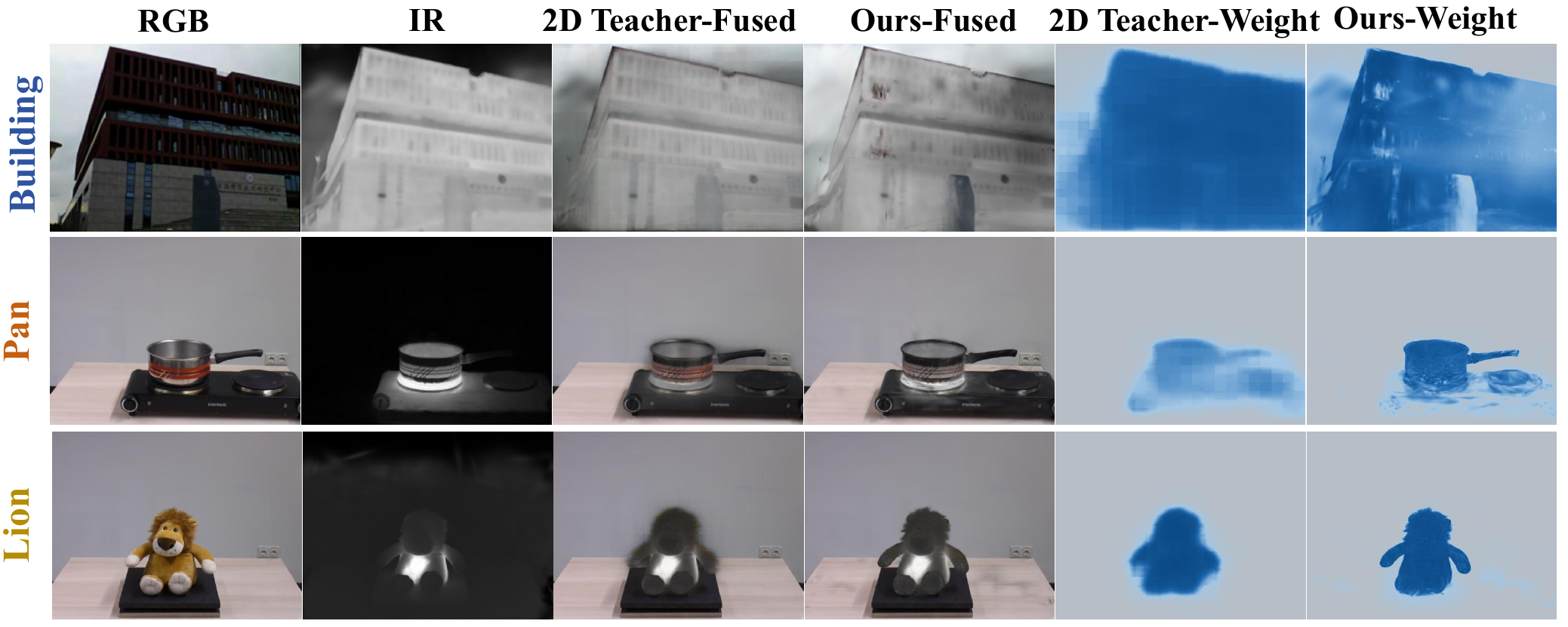}
	\vspace{-0.5cm}
	\caption{Visual validation of regional control accuracy. Teacher-weight is the CLIPSeg-predicted pixel-wise target probability maps.}
	\label{fig:native}
\end{figure}

Fig.~\ref{fig:native} visualizes LIFT's regional control, with the 2D teacher providing CLIPSeg-based modality references. LIFT accurately localizes language-specified regions and assigns corresponding modality weights. Table~\ref{tab:exp2-region} evaluates regional accuracy and cross-view consistency against the same manual annotations. Due to the contributions of multiple 3D Gaussians to individual pixels, projected weights may differ from 2D references, particularly near region boundaries. Nevertheless, LIFT achieves lower \textit{Reproj. MAE} in both target and background regions across all three scenes. Overall, LIFT enables effective language-guided regional control with improved cross-view consistency.

\begin{figure}[t]
	\centering
	\includegraphics[width=1\linewidth]{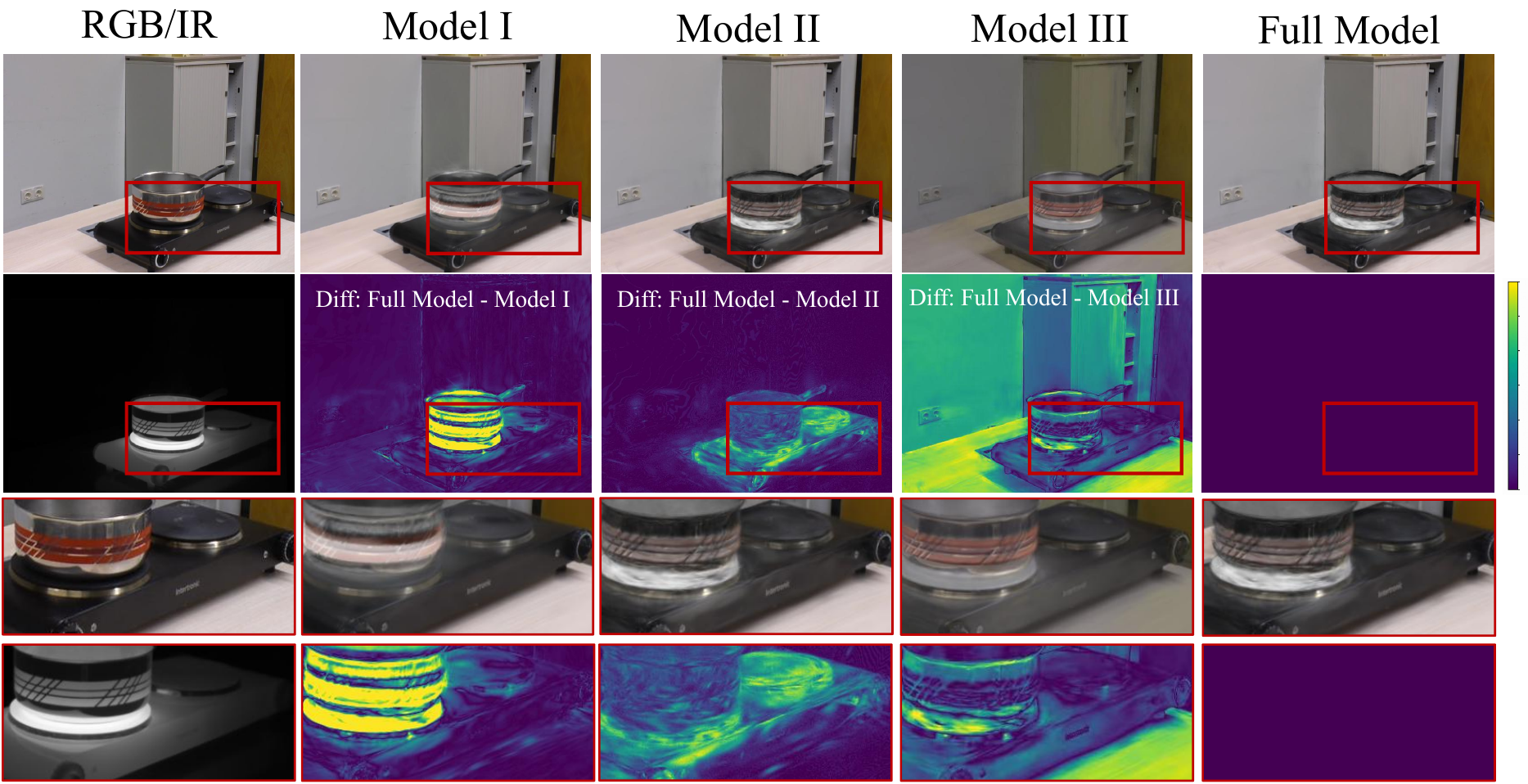}
	\vspace{-0.6cm}
	\caption{Visual Results of ablation studies on core contributions.}
	\label{fig:visual_ablation}
	\vspace{-0.2cm}
\end{figure}

\begin{table}[t]
\centering\color{black}\upshape\footnotesize
\setlength{\tabcolsep}{2pt}\renewcommand{\arraystretch}{1}
\caption{Quantitative evaluation of regional control accuracy and cross-view consistency across three scenes.}\label{tab:exp2-region}
\vspace{-0.2cm}
\begin{tabular*}{\columnwidth}{@{\extracolsep{\fill}}lcccc@{}}\toprule
&\multicolumn{2}{c}{Regional MAE $\downarrow$}&\multicolumn{2}{c}{Reproj. MAE $\downarrow$}\\
\cmidrule(lr){2-3}\cmidrule(l){4-5}
Method & Target & Backg. & Target & Backg.\\\midrule
\multicolumn{5}{@{}l}{\textbf{Building}\quad $35$ test views}\\
2D Teacher & \cellcolor{firstcolor}{\textbf{0.183}} & \cellcolor{firstcolor}{\textbf{0.168}} & 0.077 & 0.202\\
Ours & 0.267 & 0.211 & \cellcolor{firstcolor}{\textbf{0.015}} & \cellcolor{firstcolor}{\textbf{0.023}}\\
\midrule
\multicolumn{5}{@{}l}{\textbf{Lion}\quad $9$ test views}\\
2D Teacher & \cellcolor{firstcolor}{\textbf{0.125}} &\cellcolor{firstcolor}{\textbf{0.016}} & 0.093 & 0.047\\
Ours & 0.149 & 0.020 & \cellcolor{firstcolor}{\textbf{0.026}} & \cellcolor{firstcolor}{\textbf{0.022}}\\
\midrule
\multicolumn{5}{@{}l}{\textbf{Pan}\quad $11$ test views}\\
2D Teacher & 0.454 & \cellcolor{firstcolor}{\textbf{0.055}} & 0.082 & 0.049\\
Ours & \cellcolor{firstcolor}{\textbf{0.232}} & 0.075 & \cellcolor{firstcolor}{\textbf{0.033}} & \cellcolor{firstcolor}{\textbf{0.027}}\\
\bottomrule\end{tabular*}
\end{table}

\subsection{Ablation Studies}
We investigate three core contributions of our LIFT: semantic modality binding (SMB), language-conditioned Gaussian appearance (LCGA), and multi-view language-supervision (MVLS). Three ablation variants are derived against the full model:  \textbf{Model I} removes SMB scene memory to test how scene context helps bind instruction targets to RGB-IR blending weights; \textbf{Model II} fixes $A_i(u)\equiv 0.05$ in Eq.~\eqref{eq:color-main}, disabling adaptive appearance modulation in LCGA; \textbf{Model III} replaces MVLS with color images generated exclusively by CDDFuse as supervision.

Fig.~\ref{fig:visual_ablation} presents qualitative comparisons under the instruction ``\textit{ir for pan, rgb for else}''. Model I exhibits inaccurate target localization, causing infrared blending to deviate from the pan boundaries and highlighting the role of SMB in grounding language instructions to spatially consistent modality assignments. Model II shows noticeable changes in pan textures, thermal boundaries, and adjacent burner appearance, reflecting the loss of adaptive appearance modulation in LCGA. Model III exhibits weaker thermal details and contrast near the pan base, as fixed CDDFuse-generated targets provide limited instruction-dependent supervision. In contrast, our full model achieves more precise target enhancement while better preserving the color and texture of surrounding regions, demonstrating the complementary contributions of SMB, LCGA, and MVLS. Table~\ref{tab:ablation} further confirms the contributions of these components under the equal RGB-IR blending instruction, as removing any one of them causes degraded fusion quality.

\begin{figure}[t]
	\centering
	\includegraphics[width=1\linewidth]{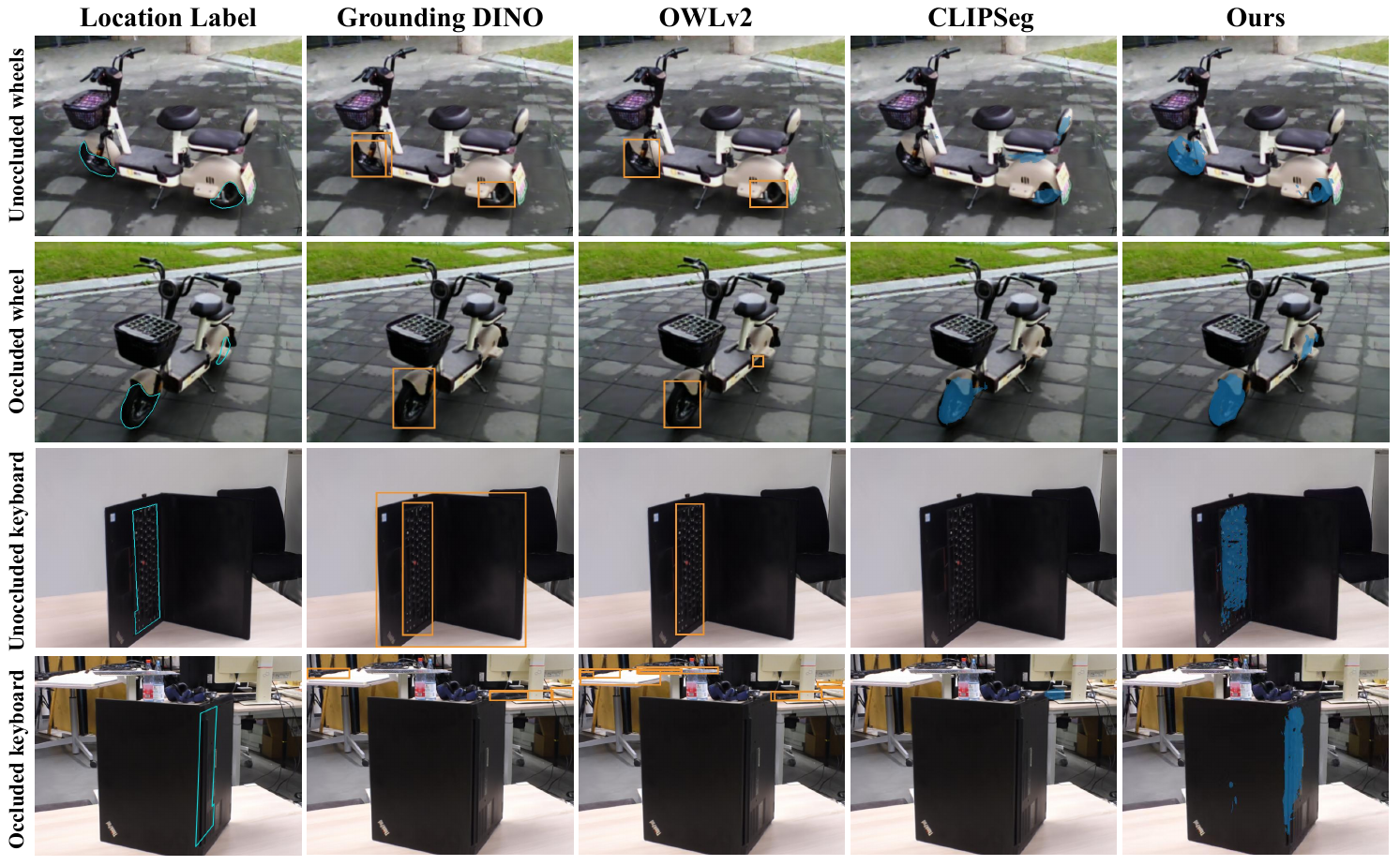}
	\vspace{-0.6cm}
	\caption{Visual localization of occluded objects.}
	\label{fig:occlusion_comparison}
\end{figure}

\subsection{Semantic Memory under Occlusion}\label{sec:occlusion_memory}
We evaluate occluded target localization on $58$ test views containing occluded targets across $15$ scenes, comparing LIFT with Grounding DINO-T~\cite{liu2024grounding}, OWLv2~\cite{minderer2023scaling}, and CLIPSeg~\cite{luddecke2022segmentation}. As shown in Fig.~\ref{fig:occlusion_comparison}, LIFT reliably localizes both partially occluded targets (\textit{e.g.}, the rear wheel) and fully occluded objects (\textit{e.g.}, the keyboard), while other methods struggle, especially under complete occlusion. Table~\ref{tab:occlusion} shows that LIFT achieves the highest Recall and IoU with competitive Precision. These results demonstrate that 3D spatial memory preserves semantic evidence across viewpoints, enabling localization beyond current visual observations.

\begin{table}[t]
	\centering
	\caption{Quantitative ablation studies on $217$ views from $15$ scenes.}\label{tab:ablation}
	\vspace{-0.3cm}
	\footnotesize
	\setlength{\tabcolsep}{1pt}
	\renewcommand{\arraystretch}{0.9}
	\begin{tabular*}{\columnwidth}{@{\extracolsep{\fill}}lcccc@{}}
		\toprule
		Variants & MSE $\downarrow$ & PSNR $\uparrow$ & CC $\uparrow$ & $N^{AB/F}\downarrow$ \\
		\midrule
		Model I    & 0.050 & 13.377 & 0.497 & 0.004 \\
		Model II     & 0.081  & 10.942 & 0.470 & \cellcolor{firstcolor}\textbf{0.001} \\
		Model III & 0.056 & 12.858 & 0.522 & 0.006 \\
		Full Model & \cellcolor{firstcolor}\textbf{0.049} & \cellcolor{firstcolor}\textbf{13.542} & \cellcolor{firstcolor}\textbf{0.524} & 0.005 \\
		\bottomrule
	\end{tabular*}
	\vspace{-0.2cm}
\end{table}

\begin{table}[t]
	\centering
	\caption{Target localization on $58$ occluded views from $15$ scenes.}\label{tab:occlusion}
	\vspace{-0.3cm}
	\small
	\setlength{\tabcolsep}{3pt}
	\renewcommand{\arraystretch}{0.9}
	\begin{tabular*}{\linewidth}{@{\extracolsep{\fill}}lccc@{}}
		\toprule
		Method & Recall $\uparrow$ & Precision $\uparrow$ & IoU $\uparrow$ \\
		\midrule
		Grounding DINO-T  & 84.213 & 16.840          & 15.594 \\
		OWLv2           & 62.494             & \cellcolor{firstcolor}\textbf{22.226} & 19.359 \\
		CLIPSeg           & 16.021             & 21.609       & 8.380 \\
		Ours     & \cellcolor{firstcolor}\textbf{89.576}    & 21.641          & \cellcolor{firstcolor}\textbf{19.616} \\
		\bottomrule
	\end{tabular*}
\end{table}

\section{Conclusion}
We present LIFT, a language-controlled 3D image fusion framework that extends image fusion to free-viewpoint 3D scenes. Through SMB, LIFT grounds language instructions in a shared 3D semantic field to establish persistent spatial modality assignments, which are further realized by LCGA through viewpoint-aware appearance modeling. We also construct the MVLS dataset to facilitate instruction-dependent fusion learning across viewpoints. Extensive experiments demonstrate that LIFT enables flexible language-driven control of multimodal scene appearances across viewpoints.

\section*{Acknowledgments}
This work was supported by the NSFC (62506268, 62676289), the Basic Research Program of Jiangsu (BK20250454), and the CPSF (GZB20250066, 2025M78150, 2026T190426). The numerical calculations in this paper have been done on the supercomputing system in the Supercomputing Center of Wuhan University. The authors declare no competing interests.

{
    \small
    \bibliographystyle{ieeenat_fullname}
    \bibliography{ref}
}


\end{document}